\documentclass[preprint,sts,11pt]{imsart}

\usepackage[top=2.5cm, bottom=2.5cm, left=2.5cm, right=2.5cm]{geometry}
\newcommand{\g}{\,\vert\,}
\newcommand{\E}{\textrm{E}}

\newcommand{\dir}{\textrm{Dir}}
\newcommand{\mult}{\textrm{Mult}}

\newcommand{\myeq}[1]{Equation \ref{eq:#1}}
\newcommand{\myfig}[1]{Figure \ref{fig:#1}}
\newcommand{\mysec}[1]{Section \ref{sec:#1}}
\newcommand{\prr}{\textrm{pR}^2}

\usepackage{natbib}
\usepackage{amsmath}
\usepackage{amssymb}
\usepackage{latexsym}
\usepackage{sectsty}
\usepackage{amsfonts}
\usepackage{epsfig}
\usepackage{url}

\newenvironment{packed_enumerate}{
  \begin{enumerate}
    \setlength{\topsep}{0pt}
    \setlength{\itemsep}{6pt}
    \setlength{\parskip}{0pt}
    \setlength{\parsep}{0pt}
}{\end{enumerate}}

\begin{document}

\begin{frontmatter}

\title{Supervised Topic Models}
\runtitle{Supervised Topic Models}

\begin{aug}
   \author{David M. Blei\ead[label=e1]{blei@cs.princeton.edu}},
   \affiliation{Princeton University}
   \author{Jon D. McAuliffe\ead[label=e2]{jon@mcauliffe.com}}
   \affiliation{University of California, Berkeley}
  \runauthor{Blei and McAuliffe}

  \address{}

\end{aug}



\begin{abstract}
  We introduce supervised latent Dirichlet allocation (sLDA), a
  statistical model of labelled documents. The model accommodates
  a variety of response types. We derive an approximate
  maximum-likelihood procedure for parameter estimation, which
  relies on variational methods to handle intractable posterior
  expectations. Prediction problems motivate this research: we
  use the fitted model to predict response values for new
  documents. We test sLDA on two real-world problems: movie
  ratings predicted from reviews, and the political tone of
  amendments in the U.S. Senate based on the amendment text. We
  illustrate the benefits of sLDA versus modern regularized
  regression, as well as versus an unsupervised LDA analysis
  followed by a separate regression.
\end{abstract}
\end{frontmatter}

\section{Introduction}

There is a growing need to analyze collections of electronic text. We have
unprecedented access to large corpora, such as government documents and
news archives, but we cannot take advantage of these collections without
being able to organize, understand, and summarize their contents. We need
new statistical models to analyze such data, and fast algorithms to compute
with those models.

The complexity of document corpora has led to considerable interest in
applying hierarchical statistical models based on what are called
\textit{topics}~\citep{Blei:2009,Blei:2003b,Griffiths:2004,Erosheva:2004}.
Formally, a topic is a probability distribution over terms in a
vocabulary. Informally, a topic represents an underlying semantic
theme; a document consisting of a large number of words might be
concisely modelled as deriving from a smaller number of topics. Such
\textit{topic models} provide useful descriptive statistics for a
collection, which facilitates tasks like browsing, searching, and
assessing document similarity.

Most topic models, such as latent Dirichlet allocation
(LDA)~\citep{Blei:2003b}, are unsupervised. Only the words in the
documents are modelled, and the goal is to infer topics that maximize
the likelihood (or the posterior probability) of the collection.  This
is compelling---with only the documents as input, one can find
patterns of words that reflect the broad themes that run through
them---and unsupervised topic modeling has many applications.
Researchers have used the topic-based decomposition of the collection
to examine interdisciplinarity~\citep{Ramage:2009}, organize large
collections of digital books~\citep{Mimno:2007}, recommend purchased
items~\citep{Marlin:2003}, and retrieve relevant documents to a
query~\citep{Wei:2006}.  Researchers have also evaluated the
interpretability of the topics directly, such as by correlation to a
thesaurus~\citep{Griffiths:2006} or by human
study~\citep{Chang:2009b}.

In this work, we focus on document collections where each document is
endowed with a response variable, external to its words, that we are
interested in predicting.  There are many examples: in a collection of
movie reviews, each document is summarized by a numerical rating; in a
collection of news articles, each document is assigned to a section of
the newspaper; in a collection of on-line scientific articles, each
document is downloaded a certain number of times.  To analyze such
collections, we develop \textit{supervised topic models}, where the
goal is to infer latent topics that are predictive of the response.
With a fitted model in hand, we can infer the topic structure of an
unlabeled document and then form a prediction of its response.

Unsupervised LDA has previously been used to construct features for
classification. The hope was that LDA topics would turn out to be
useful for categorization, since they act to reduce data
dimension~\citep{Blei:2003b, Fei-Fei:2005, Quelhas:2005}. However,
when the goal is prediction, fitting unsupervised topics may not be a
good choice.  Consider predicting a movie rating from the words in its
review.  Intuitively, good predictive topics will differentiate words
like ``excellent'', ``terrible'', and ``average,'' without regard to
genre. But topics estimated from an unsupervised model may correspond
to genres, if that is the dominant structure in the corpus.

The distinction between unsupervised and supervised topic models is
mirrored in existing dimension-reduction techniques. For example,
consider regression on unsupervised principal components versus
partial least squares and projection pursuit~\citep{Hastie:2001}---
both search for covariate linear combinations most predictive of a
response variable. Linear supervised methods have nonparametric
analogs, such as an approach based on kernel
ICA~\citep{Fukumizu:2004}.  In text analysis problems,
\cite{McCallum:2006} developed a joint topic model for words and
categories, and ~\cite{Blei:2003a} developed an LDA model to predict
caption words from images.  In chemogenomic profiling,
\cite{Flaherty:2005} proposed ``labelled LDA,'' which is also a joint
topic model, but for genes and protein function categories. These
models differ fundamentally from the model proposed here.

This paper is organized as follows. We develop the supervised latent
Dirichlet allocation model (sLDA) for document-response pairs. We
derive parameter estimation and prediction algorithms for the general
exponential family response distributions.  We show specific
algorithms for a Gaussian response and Poisson response, and suggest a
general approach to any other exponential family.  Finally, we
demonstrate sLDA on two real-world problems. First, we predict movie
ratings based on the text of the reviews. Second, we predict the
political tone of a senate amendment, based on an ideal-point analysis
of the roll call data~\citep{Clinton:2004}.  In both settings, we find
that sLDA provides more predictive power than regression on
unsupervised LDA features. The sLDA approach also improves on the
lasso~\citep{Tibshirani:1996}, a modern regularized regression
technique.

\section{Supervised latent Dirichlet allocation} \label{sec:slda}

Topic models are distributions over document collections where each
document is represented as a collection of discrete random variables
$W_{1:n}$, which are its words.  In topic models, we treat the words
of a document as arising from a set of latent topics, that is, a set
of unknown distributions over the vocabulary.  Documents in a corpus
share the same set of $K$ topics, but each document uses a mix of
topics---the topic proportions---unique to itself.  Topic models are a
relaxation of classical document mixture models, which associate each
document with a single unknown topic.  Thus they are mixed-membership
models~\citep{Erosheva:2004}.  See \cite{Griffiths:2006}
and~\cite{Blei:2009} for recent reviews.

Here we build on latent Dirichlet allocation (LDA)~\citep{Blei:2003b},
a topic model that serves as the basis for many others. In LDA, we
treat the topic proportions for a document as a draw from a Dirichlet
distribution.  We obtain the words in the document by repeatedly
choosing a topic assignment from those proportions, then drawing a
word from the corresponding topic.

In \textit{supervised latent Dirichlet allocation} (sLDA), we add to
LDA a response variable connected to each document. As mentioned,
examples of this variable include the number of stars given to a
movie, the number of times an on-line article was downloaded, or the
category of a document.  We jointly model the documents and the
responses, in order to find latent topics that will best predict the
response variables for future unlabeled documents.  SLDA uses the same
probabilistic machinery as a generalized linear model to accommodate
various types of response: unconstrained real values, real values
constrained to be positive (e.g., failure times), ordered or unordered
class labels, nonnegative integers (e.g., count data), and other
types.

\begin{figure}[t]
  \begin{center}
    \includegraphics[width=0.65\textwidth]{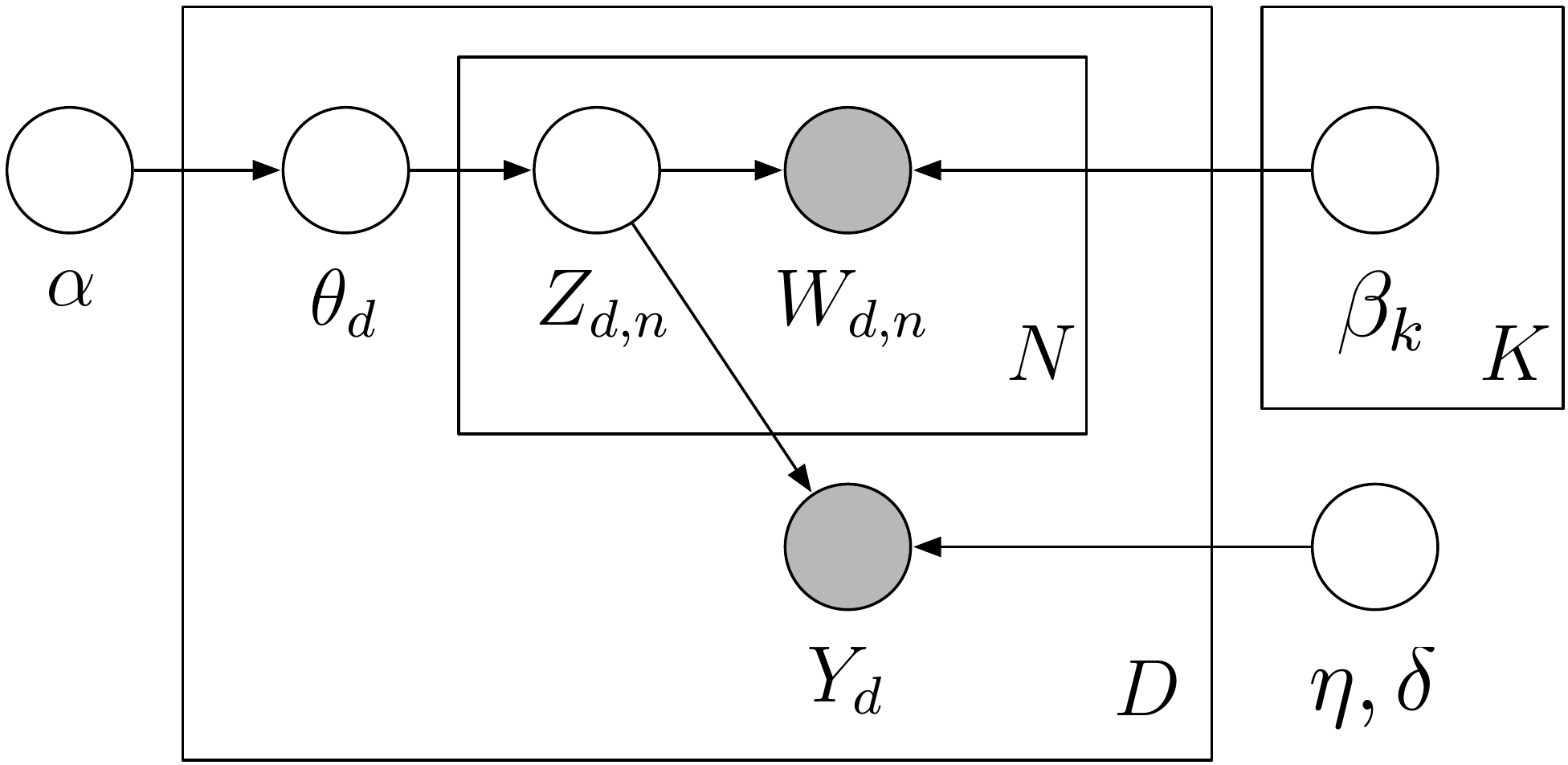}
  \end{center}
  \caption{The graphical model representation of supervised latent
    Dirichlet allocation (sLDA).  (Nodes are random variables; edges
    indicate possible dependence; a shaded node is an observed
    variable; an unshaded node is a hidden variable.)}
  \label{fig:combo}
\end{figure}

Fix for a moment the model parameters: $K$ topics $\beta_{1:K}$ (each
$\beta_k$ a vector of term probabilities), a Dirichlet parameter
$\alpha$, and response parameters $\eta$ and $\delta$.  (These
response parameters are described in detail below.)  Under the sLDA
model, each document and response arises from the following generative
process:
\begin{packed_enumerate}
  \item Draw topic proportions $\theta \g \alpha \sim
    \dir(\alpha).$
  \item For each word
    \begin{packed_enumerate}
    \item Draw topic assignment $z_n \g \theta \sim \mult(\theta).$
    \item Draw word $w_n \g z_n, \beta_{1:K} \sim
      \mult(\beta_{z_n}).$
    \end{packed_enumerate}
  \item Draw response variable $y \g z_{1:N}, \eta, \delta \sim
    \text{GLM}(\bar{z}, \eta, \delta)$, where we define
    \begin{equation}
      \label{eq:zbar}
      \bar{z} :=
      (1/N) \textstyle \sum_{n=1}^{N} z_n.
    \end{equation}
\end{packed_enumerate}
\myfig{combo} illustrates the family of probability distributions
corresponding to this generative process as a graphical model.

The distribution of the response is a generalized linear
model~\citep{McCullagh:1989},
\begin{equation}
  \label{eq:slda-glm}
  p(y \g z_{1:N}, \eta, \delta) = h(y, \delta) \exp\left\{
    \frac{(\eta^\top\bar{z}) y - A(\eta^\top \bar{z})} {\delta} \right\} \ .
\end{equation}
There are two main ingredients in a generalized linear model (GLM):
the ``random component'' and the ``systematic component.'' For the
random component, we take the distribution of the response to be an
\textit{exponential dispersion family} with natural parameter
$\eta^\top \bar{z}$ and dispersion parameter $\delta$.  For each fixed
$\delta$, \myeq{slda-glm} is an exponential family, with base measure
$h(y,\delta)$, sufficient statistic $y$, and log-normalizer
$A(\eta^\top\bar{z})$. The dispersion parameter provides additional flexibility
in modeling the variance of $y$. Note that~\myeq{slda-glm} need not be
an exponential family jointly in $(\eta^\top \bar{z}, \delta)$.

In the systematic component of the GLM, we relate the
exponential-family parameter of the random component to a linear
combination of covariates---the so-called \textit{linear
  predictor}. For sLDA, we have already introduced the linear
predictor: it is $\eta^\top \bar{z}$. The reader familiar with GLMs
will recognize that our choice of systematic component means sLDA uses
only canonical link functions. In future work, we will relax this
constraint.

The GLM framework gives us the flexibility to model any type of
response variable whose distribution can be written in exponential
dispersion form~\myeq{slda-glm}. This includes many commonly used
distributions: the normal (for real response); the binomial (for
binary response); the multinomial (for categorical response); the
Poisson and negative binomial (for count data); the gamma, Weibull,
and inverse Gaussian (for failure time data); and others. Each of
these distributions corresponds to a particular choice of
$h(y,\delta)$ and $A(\eta^\top \bar{z})$. For example, the normal
distribution corresponds to $h(y, \delta) = (1/\sqrt{2\pi\delta})
\exp\{-y^2/2\}$ and $A(\eta^\top \bar{z}) = (\eta^\top
\bar{z})^2/2$.  In this case, the usual Gaussian parameters, mean
$\mu$ and variance $\sigma^2$, are equal to $\eta^\top \bar{z}$ and
$\delta$, respectively.

What distinguishes sLDA from the usual GLM is that the covariates are
the unobserved empirical frequencies of the topics in the document.
In the generative process, these latent variables are responsible for
producing the words of the document, and thus the response and the
words are tied.  The regression coefficients on those frequencies
constitute $\eta$.  Note that a GLM usually includes an intercept
term, which amounts to adding a covariate that always equals
one. Here, such a term is redundant, because the components of
$\bar{z}$ always sum to one (see \myeq{zbar}).

By regressing the response on the empirical topic frequencies, we
treat the response as non-exchangeable with the words. The document
(i.e., words and their topic assignments) is generated first, under
full word exchangeability; then, based on the document, the response
variable is generated. In contrast, one could formulate a model in
which $y$ is regressed on the topic proportions $\theta$. This treats
the response and all the words as jointly exchangeable.  But as a
practical matter, our chosen formulation seems more sensible: the
response depends on the topic frequencies which actually occurred in
the document, rather than on the mean of the distribution generating
the topics.  Estimating this fully exchangeable model with enough
topics allows some topics to be used entirely to explain the response
variables, and others to be used to explain the word occurrences. This
degrades predictive performance, as demonstrated in~\cite{Blei:2003a}.
Put a different way, here the latent variables that govern the
response are the same latent variables that governed the words.  The
model does not have the flexibility to infer topic mass that explains
the response, without also using it to explain some of the observed
words.

\section{Computation with supervised LDA}

We need to address three computational problems to analyze data with
sLDA.  First is \textit{posterior inference}, computing the
conditional distribution of the latent variables at the document level
given its words $w_{1:N}$ and the corpus-wide model parameters.  The
posterior is thus a conditional distribution of topic proportions
$\theta$ and topic assignments $z_{1:N}$.  This distribution is not
possible to compute.  We approximate it with variational inference.

Second is \textit{parameter estimation}, estimating the Dirichlet
parameters $\alpha$, GLM parameters $\eta$ and $\delta$, and topic
multinomials $\beta_{1:K}$ from a data set of observed
document-response pairs $\{w_{d,1:N}, y_d\}_{d=1}^{D}$.  We use
maximum likelihood estimation based on variational
expectation-maximization.

The final problem is \textit{prediction}, predicting a response $y$
from a newly observed document $w_{1:N}$ and fixed values of the model
parameters.  This amounts to approximating the posterior
expectation $\E[y \g w_{1:N}, \alpha, \beta_{1:K}, \eta, \delta]$.

We treat these problems in turn for the general GLM setting of sLDA,
noting where GLM-specific quantities need to be computed or
approximated.  We then consider the special cases of a Gaussian
response and a Poisson response, for which the algorithms have an
exact form.  Finally, we suggest a general-purpose approximation
procedure for other response distributions.

\subsection{Posterior inference}
\label{sec:posterior}

Both parameter estimation and prediction hinge on posterior inference.
Given a document and response, the posterior distribution of the
latent variables is
\begin{multline}
  p(\theta, z_{1:N} \g w_{1:N}, y, \alpha, \beta_{1:K}, \eta,
  \sigma^2) = \\
  \frac{p(\theta \g \alpha) \left(\prod_{n=1}^{N} p(z_n \g \theta)
      p(w_n \g z_n, \beta_{1:K})\right) p(y \g z_{1:N}, \eta,
    \sigma^2)} {\int \! d\theta \, p(\theta \g \alpha) \sum_{z_{1:N}}
    \left(\prod_{n=1}^{N} p(z_n \g \theta) p(w_n \g z_n,
      \beta_{1:K})\right) p(y \g z_{1:N}, \eta, \sigma^2)}.
\end{multline}
The normalizing value is the marginal probability of the observed
data, i.e., the document $w_{1:N}$ and response $y$. This normalizer
is also known as the \textit{likelihood}, or the \textit{evidence}. As
with LDA, it is not efficiently computable~\citep{Blei:2003b}.  Thus,
we appeal to variational methods to approximate the posterior.

Variational methods encompass a number of types of approximation for
the normalizing value of the posterior.  For reviews,
see~\cite{Wainwright:2008} and ~\cite{Jordan:1999}.  We use mean-field
variational inference, where Jensen's inequality is used to lower
bound the normalizing value.  We let $\pi$ denote the set of model
parameters, $\pi = \{\alpha, \beta_{1:K}, \eta, \delta\}$ and
$q(\theta, z_{1:N})$ denote a \textit{variational distribution} of the
latent variables.  The lower bound is
\begin{eqnarray*}
  \log p(w_{1:N} \g \pi) &=& \log \int_{\theta} \sum_{z_{1:N}}
  p(\theta, z_{1:N}, w_{1:N} \g \pi) \\
  &=&  \log \int_{\theta} \sum_{z_{1:N}}
  p(\theta, z_{1:N}, w_{1:N} \g \pi) \frac{q(\theta,
    z_{1:N})}{q(\theta, z_{1:N})} \\
  &\geq& \E[\log p(\theta, z_{1:N}, w_{1:N} \g \pi)] - \E[ \log
  q(\theta, z_{1:N})],
\end{eqnarray*}
where all expectations are taken with respect to $q(\theta, z_{1:N})$.
This bound is called the \textit{evidence lower bound} (ELBO), which
we denote by $\mathcal{L}(\cdot)$.  The first term is the expectation
of the log of the joint probability of hidden and observed variables;
the second term is the entropy of the variational distribution,
$\textrm{H}(q) = - \E[\log q(\theta, z_{1:N})]$.  In its expanded
form, the sLDA ELBO is
\begin{eqnarray}
  \nonumber \lefteqn{\mathcal{L}(w_{1:N}, y \g \pi) =
  \E[\log
  p(\theta \g \alpha)] \ +
  \textstyle \sum_{n=1}^{N} \E[\log p(Z_n \g \theta)]} \\ & +
  \sum_{n=1}^{N} \E[\log p(w_n  \g Z_n, \beta_{1:K})]
  + \E[\log p(y \g Z_{1:N}, \eta, \delta)] +
  \text{H}(q) \ .   \label{eq:elbo}
\end{eqnarray}
This is a function of the observations $\{w_{1:N}, y\}$ and the
variational distribution.

In variational inference we first construct a parameterized family for
the variational distribution, and then fit its parameters to tighten
\myeq{elbo} as much as possible for the given observations.  The
parameterization of the variational distribution governs the expense
of optimization.

\myeq{elbo} is tight when $q(\theta, z_{1:N})$ is the posterior, but
specifying a family that contains the posterior leads to an
intractable optimization problem.  We thus specify a simpler family.
Here, we choose the fully factorized distribution,
\begin{equation}
  q(\theta, z_{1:N} \g \gamma, \phi_{1:N}) =
  q(\theta \g \gamma) \textstyle \prod_{n=1}^{N} q(z_n \g \phi_n),
  \label{eq:var-dist}
\end{equation}
where $\gamma$ is a K-dimensional Dirichlet parameter vector and each
$\phi_n$ parametrizes a categorical distribution over $K$
elements. The latent topic assignment $Z_n$ is represented as a
$K$-dimensional indicator vector; thus $\E[ Z_n ] = q(z_n) = \phi_n$.
Tightening the bound with respect to this family is equivalent to
finding its member that is closest in KL divergence to the
posterior~\citep{Wainwright:2008, Jordan:1999}.  Thus, given a
document-response pair, we maximize \myeq{elbo} with respect to
$\phi_{1:N}$ and $\gamma$ to obtain an estimate of the posterior.

Before turning to optimization, we describe how to compute the ELBO in
\myeq{elbo}.  The first three terms and the entropy of the variational
distribution are identical to the corresponding terms in the ELBO for
unsupervised LDA~\citep{Blei:2003b}.  The first three terms are
\begin{eqnarray}
  \E[\log p(\theta \g \alpha)] &=&
  \log \Gamma\left(\textstyle \sum_{i=1}^{K} \alpha_i\right)
  - \sum_{i=1}^{K} \log \Gamma(\alpha_i) +
  \sum_{i=1}^{K} (\alpha_i - 1) \E[\log \theta_i] \\
  \E[\log p(Z_n \g \theta)] &=&
  \sum_{i=1}^{K} \phi_{n,i} \E[\log \theta_i] \\
  \E[\log p(w_n \g Z_n, \beta_{1:K})] &=&
  \sum_{i=1}^{K} \phi_{n,i} \log \beta_{i, w_n}.
\end{eqnarray}
The entropy of the variational distribution is
\begin{equation}
  \text{H}(q) =
  - \sum_{n=1}^{N} \sum_{i=1}^{K} \phi_{n,i} \log \phi_{n,i}
  - \log \Gamma\left(\textstyle \sum_{i=1}^{K} \gamma_i\right)
  + \sum_{i=1}^{K} \log \Gamma(\gamma_i) -
  \sum_{i=1}^{K} (\gamma_i - 1) \E[\log \theta_i].
\end{equation}
We note that the expectation of the log of the Dirichlet random
variable is
\begin{equation}
  \textstyle \E[\log \theta_i] = \Psi(\gamma_i) -
  \Psi \left( \sum_{j=1}^{K} \gamma_j \right),
  \label{eq:e-log-theta}
\end{equation}
where $\Psi(x)$ denotes the digamma function, i.e., the first
derivative of the log of the Gamma function.

The variational objective function for sLDA differs from LDA in the
fourth term, the expected log probability of the response variable
given the latent topic assignments.  This term is
\begin{equation}
  \E[\log p(y \g Z_{1:N}, \eta, \delta)] = \log h(y, \delta) +
  {1 \over \delta} \left[ \eta^\top \left(\E\left[\bar{Z}\right] y\right)
    - \E \bigl[ A(\eta^\top\bar{Z}) \bigr] \right] \ .
\end{equation}
We see that computing the sLDA ELBO hinges on two expectations.  The
first expectation is
\begin{equation}
  \E\bigl[\bar{Z}\bigr] = \bar{\phi} = \frac{1}{N} \sum_{n=1}^{N}
  \phi_n \ ,
  \label{eq:exbar}
\end{equation}
which follows from $Z_n$ being an indicator vector.

The second expectation is $\E[A(\eta^\top \bar{Z})]$.  This is
computable in some models, such as when the response is Gaussian or
Poisson distributed, but in the general case it must be approximated.
We will address these settings in detail in \mysec{examples}.  For
now, we assume that this issue is resolvable and continue with the
procedure for approximate posterior inference.

We use block coordinate-ascent variational inference, maximizing
\myeq{elbo} with respect to each variational parameter vector in turn.
The terms that involve the variational Dirichlet $\gamma$ are
identical to those in unsupervised LDA, i.e., they do not involve the
response variable $y$.  Thus, the coordinate ascent update is as in
\cite{Blei:2003b},
\begin{equation}
  \textstyle \gamma^{\text{new}} \leftarrow \alpha + \sum_{n=1}^{N}
  \phi_n.
  \label{eq:gamma-update}
\end{equation}

The central difference between sLDA and LDA lies in the update for the
variational multinomial $\phi_n$.  Given $n \in \{1,\ldots,N\}$, the
partial derivative of the ELBO with respect to $\phi_n$ is
\begin{equation}
  \frac{\partial \mathcal{L}}{\partial \phi_n} =
  \E[\log \theta] + \E[\log p(w_n \g \beta_{1:K})] -
  \log \phi_n + 1 +
  \left(\frac{y}{N \delta}\right) \eta - \left({1 \over \delta}\right)
  {\partial \over \partial \phi_n}
  \left\{ \E\bigl[ A(\eta^\top \bar{Z}) \bigr] \right\} .
  \label{eq:phi-gradient}
\end{equation}
Optimizing with respect to the variational multinomial depends on the
form of ${\partial \over \partial \phi_n} \left\{ \E\bigl[ A(\eta^\top
  \bar{Z}) \bigr] \right\}$.  In some cases, such as a Gaussian or
Poisson response, we obtain exact coordinate updates.  In other cases,
we require gradient based optimization methods.  Again, we postpone
these details to \mysec{examples}.

Variational inference proceeds by iteratively updating the variational
parameters $\{\gamma, \phi_{1:N}\}$ according to \myeq{gamma-update}
and the derivative in \myeq{phi-gradient}.  This finds a local optimum
of the ELBO in \myeq{elbo}.  The resulting variational distribution
$q$ is used as a proxy for the posterior.

\subsection{Parameter estimation}

The parameters of sLDA are the $K$ topics $\beta_{1:K}$, the Dirichlet
hyperparameters $\alpha$, the GLM coefficients $\eta$, and the GLM
dispersion parameter $\delta$.  We fit these parameters with
variational expectation maximization (EM), an approximate form of
expectation maximization where the expectation is taken with respect
to a variational distribution.  As in the usual EM algorithm,
maximization proceeds by maximum likelihood estimation under expected
sufficient statistics~\citep{Dempster:1977}.

Our data are a corpus of document-response pairs ${\cal D} = \{w_{d,
  1:N}, y_d\}_{d=1}^{D}$.  Variational EM is an optimization of the
corpus-level lower bound on the log likelihood of the data, which is
the sum of per-document ELBOs.  As we are now considering a collection
of document-response pairs, in this section we add document indexes to
the previous section's quantities, so response variable $y$ becomes
$y_d$, empirical topic assignment frequencies $\bar{Z}$ becomes
$\bar{Z}_d$, and so on.  Notice each document is endowed with its own
variational distribution.  Expectations are taken with respect to that
document-specific variational distribution $q_d(z_{1:N}, \theta)$,
\begin{equation}
  \label{eq:corpus-elbo}
  {\cal L}(\alpha, \beta_{1:K}, \eta, \delta ; {\cal D}) =
  \sum_{d=1}^{D} \E_d[\log p(\theta_d, z_{d,1:N}, w_{d,1:N}, y_d)] +
  \textrm{H}(q_d).
\end{equation}

In the expectation step (E-step), we estimate the approximate
posterior distribution for each document-response pair using the
variational inference algorithm described above.  In the maximization
step (M-step), we maximize the corpus-level ELBO with respect to the
model parameters. Variational EM finds a local optimum of
\myeq{corpus-elbo} by iterating between these steps.  The M-step
updates are described below.

\paragraph{Estimating the topics.} The M-step updates of the topics
$\beta_{1:K}$ are the same as for unsupervised LDA, where the probability
of a word under a topic is proportional to the expected number of times
that it was assigned to that topic~\citep{Blei:2003b},
\begin{equation}
  \hat{\beta}_{k,w}^{\text{new}} \propto
  \sum_{d=1}^{D} \sum_{n=1}^{N} 1(w_{d,n} = w) \phi_{d,n}^k .
  \label{eq:topic-m-step}
\end{equation}
Proportionality means that each $\hat{\beta}_k^{\text{new}}$ is
normalized to sum to one.  We note that in a fully Bayesian sLDA
model, one can place a symmetric Dirichlet prior on the topics and use
variational Bayes to approximate their posterior.  This adds no
additional complexity to the algorithm~\citep{Bishop:2006}.

\paragraph{Estimating the GLM parameters.}  The GLM parameters are the
coefficients $\eta$ and dispersion parameter $\delta$.  For the
corpus-level ELBO, the gradient with respect to GLM coefficients
$\eta$ is
\begin{eqnarray}
  {\partial {\cal L} \over \partial \eta} &=&
  {\partial \over \partial \eta}
  \left( {1 \over \delta} \right)
  \sum_{d=1}^D \left\{
    \eta^\top \E \bigl[\bar{Z}_d\bigr] y_d -
    \E\bigl[A(\eta^\top \bar{Z}_d)\bigr]
  \right\} \\
  &=&
  \left( {1 \over \delta} \right)
  \left\{
    \sum_{d=1}^D \bar{\phi}_d y_d
    -
    \sum_{d=1}^D \E_d \bigl[
    \mu(\eta^\top \bar{Z}_d) \bar{Z}_d  \bigr]
  \right\},
  \label{eq:glm-eta-gradient}
\end{eqnarray}
where $\mu(\cdot) = \E_{\text{GLM}}[ Y \g \cdot]$.  The appearance of
this expectation follows from properties of the cumulant generating
function $A(\eta^\top \bar{z})$ ~\citep{Brown:1986}.  This GLM mean
response is a known function of $\eta^\top \bar{z}_d$ in all standard
cases. However, $\E[ \mu(\eta^\top \bar{Z}_d)\bar{Z}_d ]$ has an exact
solution only in some cases, such as Gaussian or Poisson response. In
other cases, we approximate the expectation.  (See \mysec{examples}.)

The derivative with respect to $\delta$, evaluated at
$\hat{\eta}_{\text{new}}$, is
\begin{equation}
  \left\{
    \sum_{d=1}^D
    \frac{\partial h(y_d, \delta) / \partial \delta}{h(y_d, \delta)}
  \right\} +
    \left( {1 \over \delta} \right)
  \left\{
    \sum_{d=1}^D
    \left[\hat{\eta}_\text{new}^\top \left(\E\left[\bar{Z_d}\right] y_d\right)
      - \E \bigl[A(\hat{\eta}_\text{new}^\top\bar{Z_d}) \bigr] \right]
  \right\} \ .
  \label{eq:delta-deriv}
\end{equation}
\myeq{delta-deriv} can be computed given that the rightmost summation
has been evaluated, exactly or approximately, while optimizing the
coefficients $\eta$. Depending on $h(y,\delta)$ and its partial
derivative with respect to $\delta$, we obtain
$\hat{\delta}_{\text{new}}$ either in closed form or via
one-dimensional numerical optimization.

\paragraph{Estimating the Dirichlet parameter.}  While we fix the
Dirichlet parameter $\alpha$ to $1/K$ in \mysec{results}, other
applications of topic modeling fit this
parameter~\citep{Blei:2003b,Wallach:2009}.  Estimating the Dirichlet
parameters follows the standard procedure for estimating a Dirichlet
distribution~\citep{Ronning:1989}.  In a fully-observed setting, the
sufficient statistics of the Dirichlet are the logs of the observed
simplex vectors.  Here, they are the expected logs of the topic
proportions, see \myeq{e-log-theta}.

\subsection{Prediction} \label{sec:pred}

Our focus in applying sLDA is prediction.  Given a new document
$w_{1:N}$ and a fitted model $\{\alpha, \beta_{1:K}, \eta, \delta\}$,
we want to compute the expected response value,
\begin{equation}
  \E[Y \g w_{1:N}, \alpha, \beta_{1:K}, \eta, \sigma^2] =
  \E\bigl[\mu(\eta^\top \bar{Z}) \g w_{1:N}, \alpha, \beta_{1:K}\bigr]
  . \label{eq:predict}
\end{equation}

To perform prediction, we approximate the posterior mean of $\bar{Z}$
using variational inference.  This is the same procedure as in
\mysec{posterior}, but here the terms depending on the response $y$
are removed from the ELBO in \myeq{elbo}.  Thus, with the variational
distribution taking the same form as \myeq{var-dist}, we implement the
following coordinate ascent updates for the variational parameters,
\begin{eqnarray}
  \gamma^\textrm{new} &=& \textstyle \alpha + \sum_{n=1}^{N} \phi_n
  \label{eq:var-lda-gamma}\\
  \phi_{n}^\textrm{new} &\propto& \exp\{\E_q[\log \theta] + \log
  \beta_{w_n}\}
  \label{eq:var-lda-phi},
\end{eqnarray}
where the log of a vector is a vector of the log of each of its
components, $\beta_{w_n}$ is the vector of $p(w_n \g \beta_k)$ for
each $k \in \{1, \ldots, K\}$, and again proportionality means that
the vector is normalized to sum to one.  Note in this section we
distinguish between expectations taken with respect to the model and
those taken with respect to the variational distribution.  (In other
sections all expectations are taken with respect to the variational
distribution.)

This coordinate ascent algorithm is identical to variational inference
for unsupervised LDA: since we averaged the response variable out of
the right-hand side in~\myeq{predict}, what remains is the standard
unsupervised LDA model for $Z_{1:N}$ and $\theta$~\citep{Blei:2003b}.
Notice this algorithm does not depend on the particular response type.

Thus, given a new document, we first compute $q(\theta, z_{1:N})$, the
variational posterior distribution of the latent variables $\theta$
and $Z_n$. We then estimate the response with
\begin{equation}
  \E[Y \g w_{1:N}, \alpha, \beta_{1:K}, \eta, \sigma^2] \approx
  \E_q\bigl[\mu(\eta^\top \bar{Z})\bigr]
\end{equation}
As with parameter estimation, this depends on being able to compute or
approximate $\E_q[ \mu(\eta^\top \bar{Z})]$.

\subsection{Examples} \label{sec:examples}

The previous three sections have outlined the general computational
strategy for sLDA: a procedure for approximating the posterior
distribution of the topic proportions $\theta$ and topic assignments
$Z_{1:N}$ on a per-document/response basis, a maximum likelihood
procedure for fitting the topics $\beta_{1:K}$ and GLM parameters
$\{\eta, \delta\}$ using variational EM, and a procedure for
predicting new responses from observed documents using an approximate
expectation based on a variational posterior.

Our derivation has remained free of any specific assumptions about the
form of the response distribution.  We now focus on sLDA for specific
response distributions---the Gaussian and Poisson---and suggest a
general approximation technique for handling other responses within
the GLM framework.  The response distribution blocked our derivation
at three points: the computation of $\E[A(\eta^\top \bar{Z})]$ in the
per-document ELBO of \myeq{elbo}, the computation of ${\partial
  \over \partial \phi_n} \left\{ \E\bigl[ A(\eta^\top \bar{Z}) \bigr]
\right\}$ in \myeq{phi-gradient} and corresponding update for the
variational multinomial $\phi_n$, and the computation of
$\E_q[\mu(\eta^\top \bar{Z}_d)\bar{Z}_d]$ for fitting the GLM
parameters in a variational EM algorithm.  We note that other aspects
of working with sLDA are general to any type of exponential family
response.

\paragraph{Gaussian response}
When the response is Gaussian, the dispersed exponential family form
can be seen as follows:
\begin{eqnarray}
  \label{eq:gauss-response}
  p(y \g \bar{z}, \eta, \delta) &=&
  \frac{1}{\sqrt{2 \pi \delta}}
  \exp\left\{-
    \frac{\bigl(y - \eta^\top
      \bar{z}\bigr)^2}{2\delta}\right\}. \\
  &=& \frac{1}{\sqrt{2 \pi \delta}}
  \exp\left\{
    \frac{-y^2/2 +
      y \eta^\top \bar{z} -
      (\eta^\top \bar{z} \bar{z}^\top \eta)/2}{\delta}
    \right\}\ .
\end{eqnarray}
Here the natural parameter $\eta^\top \bar{z}$ and mean parameter are
identical.  Thus, $\mu(\eta^\top \bar{z}) = \eta^\top \bar{z}$.  The
usual variance parameter $\sigma^2$ is equal to $\delta$.  Moreover,
notice that
\begin{eqnarray}
  h(y, \delta) &=&  \frac{1}{\sqrt{2 \pi \delta}} \exp\{-y^2/2\} \\
  \label{eq:gauss-h}
  A(\eta^\top \bar{z}) &=& (\eta^\top \bar{z} \bar{z}^\top
  \eta)/2 \label{eq:gauss-a}\ .
\end{eqnarray}

We first specify the variational inference algorithm.  This requires
computing the expectation of the log normalizer in \myeq{gauss-a},
which hinges on $\E\bigl[\bar{Z}\bar{Z}^\top\bigr]$,
\begin{eqnarray}
  \E\bigl[\bar{Z}\bar{Z}^\top\bigr]
  &=& \frac{1}{N^2} \textstyle
  \sum_{n=1}^{N} \sum_{m=1}^{N} \E\bigl[Z_n Z_m^\top\bigr] \\
  &=& \frac{1}{N^2} \left( \textstyle \sum_{n=1}^{N} \sum_{m \neq n}
    \phi_n \phi_m^\top + \sum_{n=1}^{N} \text{diag}\{\phi_n\}\right).
  \label{eq:exxt}
\end{eqnarray}
To see~\myeq{exxt}, notice that for $m \neq n$, $\E[Z_n Z_m^\top] =
\E[Z_n]\E[Z_m]^\top = \phi_n \phi_m^\top$ because the variational
distribution is fully factorized. On the other hand, $\E[Z_n Z_n^\top]
= \text{diag}(\E[Z_n]) = \text{diag}(\phi_n)$ because $Z_n$ is an
indicator vector.

To take the derivative of the expected log normalizer, we consider it
as a function of a single variational parameter $\phi_j$.  Define
$\phi_{-j} := \sum_{n \neq j} \phi_n$.  Expressed as a function of
$\phi_j$, $\E\bigl[A(\eta^\top \bar{Z})\bigr]$ is
\begin{eqnarray}
  f(\phi_j) &=& \left({1 \over 2 N^2
      }\right) \eta^\top \left[ \phi_j \phi_{-j}^\top + \phi_{-j}
    \phi_j^\top + \text{diag}\{\phi_j\} \right] \eta + \text{const} \\
  &=& \left({1 \over 2 N^2 }\right) \left[ 2 \bigl(\eta^\top
    \phi_{-j}\bigr) \eta^\top \phi_j + (\eta \circ \eta)^\top \phi_j
  \right] + \textrm{const} \ .
\end{eqnarray}
Thus the gradient is
\begin{equation}
  \frac{\partial}{\partial \phi_j} \E\bigl[A(\eta^\top \bar{Z})\bigr] =
  {1 \over 2 N^2} \left[ 2 \bigl(\eta^\top \phi_{-j}\bigr) \eta +
    (\eta \circ \eta) \right] \ .
  \label{eq:phi-grad-gaussian}
\end{equation}
Substituting this gradient into \myeq{phi-gradient} yields an exact
coordinate update for the variational multinomial $\phi_j$,
\begin{equation}
  \label{eq:phi-update}
  \phi_j^\text{new} \propto \exp\biggl\{
  \E[\log \theta] + \E[\log p(w_j \g \beta_{1:K})] \ + \\
  \left(\frac{y}{N\delta}\right) \eta -
  {1 \over 2 N^2 \delta} \left[ 2 \bigl(\eta^\top \phi_{-j}\bigr) \eta +
    (\eta \circ \eta) \right] \biggr\} \ .
\end{equation}
Exponentiating a vector means forming the vector of exponentials. The
proportionality symbol means the components of $\phi_j^{\text{new}}$ are
computed according to~\myeq{phi-update}, then normalized to sum to one.
Note that $\E[\log \theta_i]$ is given in \myeq{e-log-theta}.

Examining this update in the Gaussian setting uncovers further
intuitions about the difference between sLDA and LDA.  As in LDA, the
$j$th word's variational distribution over topics depends on the
word's topic probabilities under the actual model (determined by
$\beta_{1:K}$).  But $w_j$'s variational distribution, and those of
all other words, affect the probability of the response.  To see this,
consider the expectation of $\log p(y \g \bar{z}, \eta, \delta)$ of
\myeq{gauss-response}, which is is a term in the document-level ELBO
of \myeq{elbo}.  Notice that variational distribution $q(z_n)$ plays a
role in the expected residual sum of squares $\E[(y - \eta^\top
\bar{Z})^2]$.  The end result is that the update~\myeq{phi-update}
also encourages the corresponding variational parameter $\phi_j$ to
decrease this expected residual sum of squares.

Further, the update in~\myeq{phi-update} depends on the variational
parameters $\phi_{-j}$ of all other words. Unlike LDA, the $\phi_j$
cannot be updated in parallel.  Distinct occurrences of the same term
must be treated separately.

We now turn to parameter estimation for the Gaussian response sLDA
model, i.e., the M-step updates for the parameters $\eta$ and
$\delta$.  Define $y:=y_{1:D}$ as the vector of response values across
documents and let $X$ be the $D \times K$ design matrix, whose rows
are the vectors $\bar{Z}_d$.  Setting to zero the $\eta$ gradient of
the corpus-level ELBO from \myeq{glm-eta-gradient}, we arrive at an
expected-value version of the normal equations:
\begin{equation}
  \E\bigl[ X^\top X \bigr] \eta = \E[X]^\top y
  \qquad \Rightarrow \qquad \hat{\eta}_{\text{new}}  \leftarrow \left(
    \E\bigl[ X^\top X \bigr] \right)^{-1} \E[X]^\top y \
  . \label{eq:etanew}
\end{equation}
Here the expectation is over the matrix $X$, using the variational
distribution parameters chosen in the previous E-step.  Note that the
$d$th row of $\E[X]$ is just $\E[\bar{Z}_d]$ from~\myeq{exbar}. Also,
\begin{equation}
  \E\bigl[ X^\top X \bigr] = \sum_d \E\bigl[\bar{Z}_d \bar{Z}_d^\top\bigr],
\end{equation}
with each term having a fixed value from the previous E-step as well,
given by~\myeq{exxt}.

We now apply the first-order condition for $\delta$ of
\myeq{delta-deriv}.  The partial derivative needed is
\begin{equation}
  \frac{\partial h(y_d, \delta) / \partial \delta}{h(y_d, \delta)} =
  -\frac{1}{2 \delta},
\end{equation}
which can be seen from \myeq{gauss-h}.  Using this derivative and the
definition of $\hat{\eta}_{\text{new}}$ in \myeq{delta-deriv}, we
obtain
\begin{equation}
  \hat{\delta}_{\text{new}} \leftarrow {1 \over D}
  \left\{ y^\top y - y^\top \E[X] \left( \E\bigl[ X^\top X \bigr] \right)^{-1}
    \E[X]^\top  y \right\} \ . \label{eq:rss}
\end{equation}
It is tempting to try a further simplification of~\myeq{rss}: in an
ordinary least squares (OLS) regression of $y$ on the columns of
$\E[X]$, the analog of~\myeq{rss} would just equal $1/D$ times the sum
of the squared residuals $(y - \E[X] \hat{\eta}_{\text{new}})$. However,
that identity does not hold here, because the inverted matrix
in~\myeq{rss} is $\E[X^\top X]$, rather than $\E[X]^\top\E[X]$. This
illustrates that the $\eta$ update~\myeq{etanew} is not just OLS
regression of $y$ on $\E[X]$.

Finally, we form predictions just as in Section~\ref{sec:pred}.  Since
the mapping from the natural parameter to the mean parameter is the
identity function,
\begin{equation}
  \E[Y \g w_{1:N}, \alpha, \beta_{1:K}, \eta, \delta] \approx
  \eta^\top \E\bigl[\bar{Z}\bigr].
\end{equation}
Again, the expectation is taken with respect to variational inference
in the unsupervised setting (\myeq{var-lda-gamma} and
\myeq{var-lda-phi}).

\paragraph{Poisson response}

An overdispersed Poisson response provides a natural generalized
linear model formulation for count data.  Given mean parameter
$\lambda$, the overdispersed Poisson has density
\begin{equation}
  p(y \g \lambda, \delta) = \frac{1}{y!} \lambda^{y/\delta}
    \exp\{-\lambda/\delta \}.
  \end{equation}
This can be put in the overdispersed exponential family form
\begin{equation}
  p(y \g \lambda) = \frac{1}{y!}\exp\left\{\frac{y \log \lambda -
      \lambda}{\delta}\right\}.
\end{equation}
In the GLM, the natural parameter $\log \lambda = \eta^\top \bar{z}$,
$h(y, \delta) = 1/y!$, and
\begin{equation}
  A(\eta^\top \bar{z}) = \mu(\eta^\top \bar{z}) = \exp\bigl\{\eta^\top
  \bar{z} \bigr\} \ .
\end{equation}

We first compute the expectation of the log normalizer
\begin{eqnarray}
  \E\left[A(\eta^\top \bar{Z})\right] &=& \E\left[\exp\bigl\{\textstyle (1/N) \sum_{n=1}^{N}
    \eta^\top Z_n\bigr\}\right] \\
  &=& \prod_{n=1}^{N} \E\bigl[\exp\bigl\{(1/N) \eta^\top
  Z_n\bigr\}\bigr],
  \label{eq:poisson-expect-lognorm}
\end{eqnarray}
where
\begin{eqnarray}
  \E\bigl[\exp\bigl\{(1/N) \eta^\top Z_n\bigr\}\bigr] &=& \textstyle
  \sum_{i=1}^{K} \phi_{n,i} \exp\{\eta_i/N\} + (1-\phi_{n,i}) \\
  &=& \textstyle
  K - 1 + \sum_{i=1}^{K} \phi_{n,i} \exp\{\eta_i/N\}
  \ .
\end{eqnarray}
The product in \myeq{poisson-expect-lognorm} also provides
$\E[\mu(\eta^\top \bar{Z})]$, which is needed for prediction in
\myeq{predict}.  Denote this product by $C$ and let $C_{-n}$ denote
the same product, but over all indices except for the $n$th.  The
derivative of the expected log normalizer needed to update $\phi_{n}$
is
\begin{equation}
  \frac{\partial}{\partial \phi_n} \E\bigl[A(\eta^\top \bar{Z})\bigr] =
  C_{-n} \exp\{\eta/N\} \ .
\end{equation}
As for the Gaussian response, this permits an exact update of the
variational multinomial.

In the derivative of the corpus-level ELBO with respect to the
coefficients $\eta$ (\myeq{glm-eta-gradient}), we need to compute the
expected mean parameter times $\bar{Z}$,
\begin{eqnarray}
  \E[\mu(\eta^\top \bar{Z}) \bar{Z}] &=&
  \frac{1}{N} \sum_{n=1}^{N} \E[\mu(\eta^\top \bar{Z}) Z_n] \\
  &=& \frac{\exp\{\eta/N\}}{N} \sum_{n=1}^{N} C_{-n}  \phi_{n}.
\end{eqnarray}

We turn to the M-step.  We cannot find an exact M-step update for the
GLM coefficients, but we can compute the gradient for use in a convex
optimization algorithm.  The gradient of the corpus-level ELBO is
\begin{equation}
  {\partial {\cal L} \over \partial \eta} =
  \frac{1}{\delta}\left(\sum_d \E_d[\bar{Z_d}] y_d - \sum_d
    \frac{\exp\{\eta/N_d\}}{N_d} \sum_n C_{d,-n} \phi_{d,n} \right).
\end{equation}
The derivative of $h(y, \delta)$ with respect to $\delta$ is zero.
Thus the dispersion parameter M-step is exact,
\begin{equation}
  \hat{\delta}_{\textrm{new}} \rightarrow
  \frac
  {\sum_{d} \hat{\eta}_{\textrm{new}}^\top \E\bigl[\bar{Z}_d\bigr] y_d}
  {\sum_{d} \E\bigl[A(\hat{\eta}_{\textrm{new}}^\top \bar{Z}_d)\bigr]}
\end{equation}

As for the general case, Poisson sLDA prediction requires computing
$\E\bigl[\mu(\eta^\top \bar{Z})\bigr]$.  Since $\mu(\cdot)$ and
$A(\cdot)$ are identical, this is given in
\myeq{poisson-expect-lognorm}.

\paragraph{Exponential family response via the delta method} With a
general exponential family response one can use the multivariate delta
method for moments to approximate difficult
expectations~\citep{Bickel:2007}, a method which is effective in
variational approximations~\citep{Braun:2010}.  In other work,
\cite{Chang:2010} use this method with logistic regression to adapt
sLDA to network analysis; \cite{Wang:2009a} use this method with
multinomial regression for image classification.  With the
multivariate delta method, one can embed any generalized linear model
into sLDA.


\section{Empirical study}
\label{sec:results}

\begin{figure}[t]
  \begin{center}
    \includegraphics[width=\textwidth]{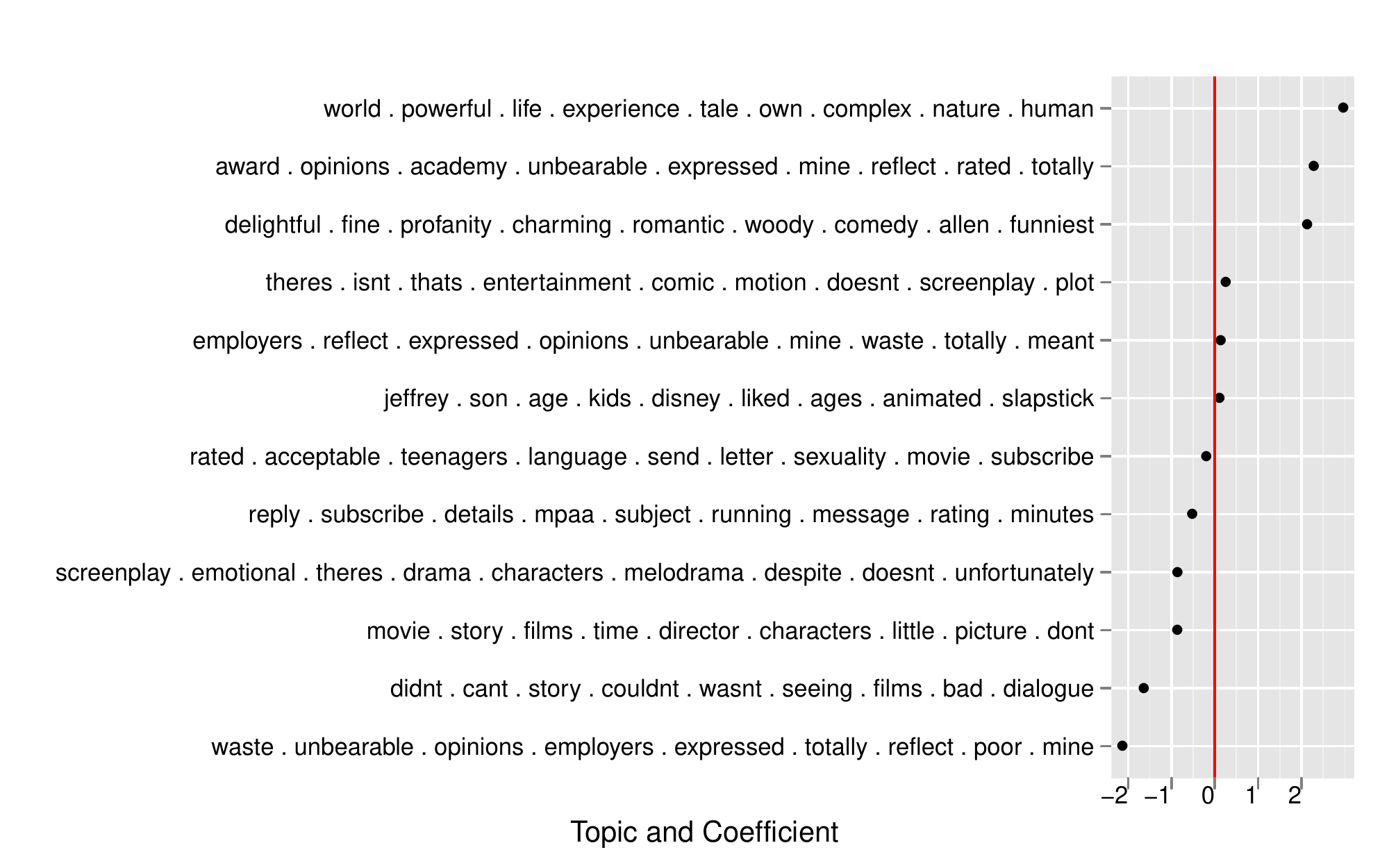}
  \end{center}
  \caption{A 12-topic sLDA model fit to the movie review data.
    \label{fig:review-model}}
\end{figure}

\begin{figure}[t]
  \begin{center}
    \includegraphics[width=\textwidth]{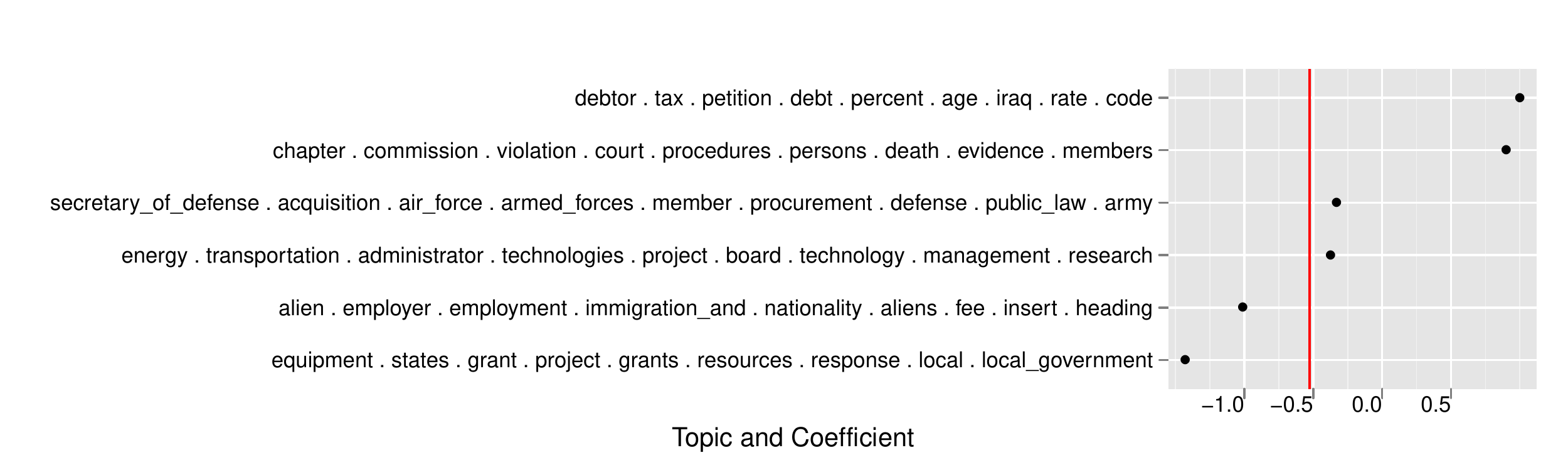}
  \end{center}
  \caption{A 5-topic sLDA model fit to the 109th U.S. Senate data.
    \label{fig:s109-model}}
\end{figure}

\begin{figure}[t]
  \begin{center}
    \includegraphics[angle=90, scale=0.7]{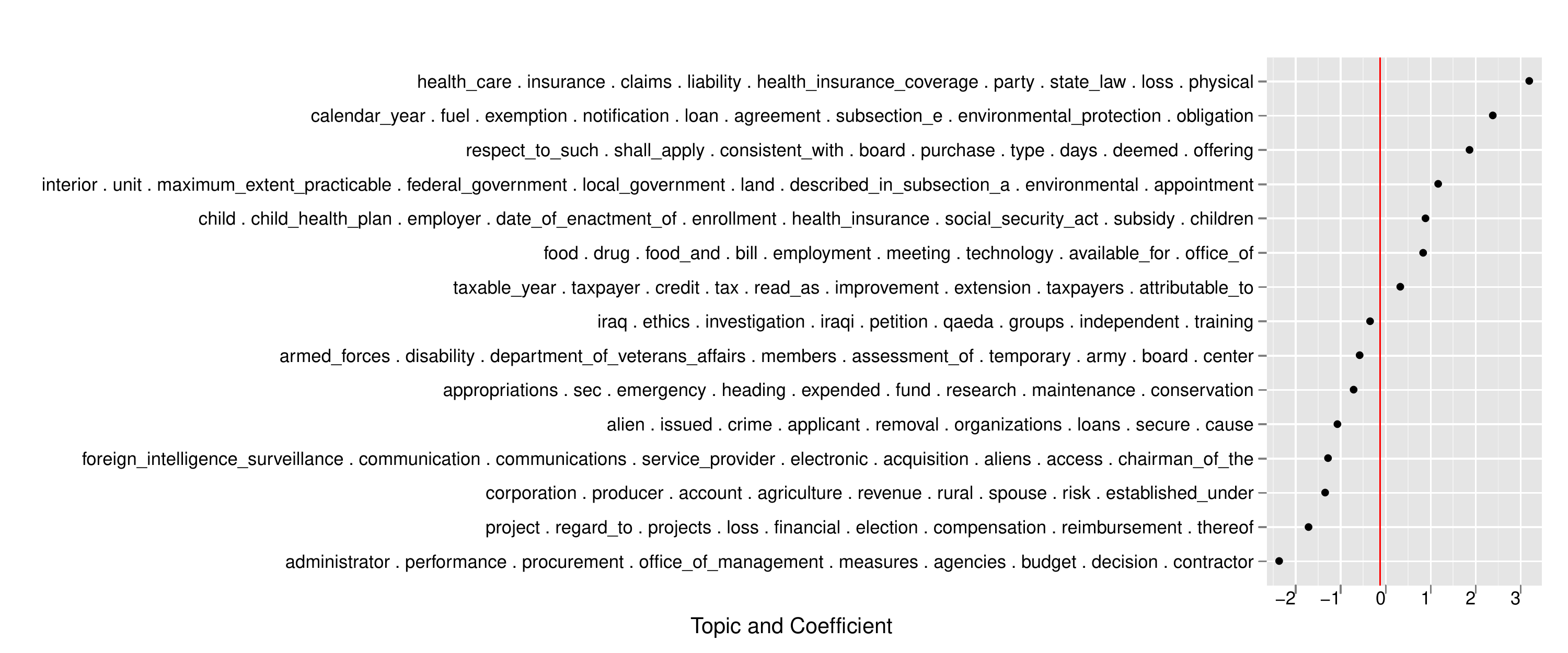}
  \end{center}
  \caption{A 15-topic sLDA model fit to the 110th U.S. Senate data.
    \label{fig:s110-model}}
\end{figure}

We studied sLDA on two prediction problems.  First, we consider the
``sentiment analysis'' of newspaper movie reviews.  We use the
publicly available data introduced in~\cite{Pang:2005}, which contains
movie reviews paired with the number of stars given. While
\cite{Pang:2005} treat this as a classification problem, we treat it
as a regression problem.

Analyzing document data requires choosing an appropriate vocabulary on
which to estimate the topics.  In topic modeling research, one
typically removes very common words, which cannot help discriminate
between documents, and very rare words, which are unlikely to be of
predictive importance.  We select the vocabulary by removing words
that occur in more than 25\% of the documents and words that occur in
fewer than 5 documents.  This yielded a vocabulary of 2180 words, a
corpus of 5006 documents, and 908,000 observations.  For more on
selecting a vocabulary in topic models, see~\cite{Blei:2009}.

Second, we studied the texts of amendments from the 109th and 110th
Senates.  Here the response variables are the discrimination
parameters from an ideal point analysis of the
votes~\citep{Clinton:2004}.  Ideal point analysis of voting data is
used in quantitative political science to map senators to a
real-valued point on a political spectrum.  Ideal point models posit
that the votes of each senator $j$ are summarized with a real-valued
latent variable $x_j$. The senator's vote on an issue $i$, which is
the binary variable $y_{ij}$, is determined by a probit model,
\begin{equation*}
  y_{ij} \sim \textrm{Probit}(x_j \beta_i + \alpha_i).
\end{equation*}
Thus, each issue is connected to two parameters.  For our analysis, we
are most interested in the ``issue discrimination'' $\beta_i$.  When
$\beta_i$ and $x_j$ have the same sign, the senator $j$ is more likely
to vote in favor of the issue $i$.  Just as $x_j$ can be interpreted
as a senator's point on the political spectrum, $\beta_i$ can be
interpreted as an issue's place on that spectrum.  (The intercept term
$\alpha_i$ is called the ``difficulty'' and allows for bias in the
votes, regardless of the ideal points of the senators.)

We connected the results of an ideal point analysis to texts about the
votes.  In particular, we study the amendments considered by the 109th
and 110th U.S. Senates.\footnote{We collected our data---the amendment
  texts and the voting records---from the open government web-site
  www.govtrack.com.}  As a preprocessing step, we estimated the
discrimination parameters $\beta_i$ for each amendment, based on the
voting record.  Each discrimination $\beta_i$ is used as the response
variable for its amendment text.  We study the question: How well we
can predict the political tone of an amendment based only on its text?

We pre-processed the Senate text in the same way as the reviews data.
To obtain the response variables, we used the implementation of ideal
point analysis in Simon Jackman's Political Science Computational
Laboratory R package.  For the 109th Senate, this yielded 288
amendments, a vocabulary of 2084 words, and 62,000 observations.  For
the 110th Senate, this yielded 213 amendments, a vocabulary of 1653
words, and 63,000 observations.

For both reviews and senate amendments, we transformed the response to
approximate normality by taking logs. This makes the data amenable to
the continuous-response model of \mysec{slda}; for these two problems,
generalized linear modeling turned out to be unnecessary.  We
initialized $\beta_{1:K}$ to randomly perturbed uniform topics,
$\sigma^2$ to the sample variance of the response, and $\eta$ to a
grid on $[-1, 1]$ in increments of $2/K$. We ran EM until the relative
change in the corpus-level likelihood bound was less than 0.01\%. In
the E-step, we ran coordinate-ascent variational inference for each
document until the relative change in the per-document ELBO was less
than 0.01\%.

We can examine the patterns of words found by models fit to these
data.  In \myfig{review-model} we illustrate a 12-topic sLDA model fit
to the movie review data.  Each topic is plotted as a list of its most
likely words, and each is attached to its estimated coefficient in the
linear model.  Words like ``powerful'' and ``complex'' are in a topic
with a high positive coefficient; words like ``waste'' and
``unbearable'' are in a topic with high negative coefficient.  In
\myfig{s109-model} and \myfig{s110-model} we illustrate sLDA models
fit to the Senate amendment data.  These models are harder to
interpret than the models fit to movie review data, as the response
variable is likely less governed by the text.  (Much more goes into a
Senator's decision of how to vote.)  Nonetheless, some patterns are
worth noting.  The health care amendments in the 110th Senate were a
distinctly right wing issue; grants and immigration in the 109th
Senate were a left wing issue.

We assessed the quality of the predictions using five fold
cross-validation.  We measured error in two ways.  First, we measured
correlation between the out-of-fold predictions with the out-of-fold
response variables.  Second, we measured ``predictive R$^2$.'' We
defined this quantity as the fraction of variability in the
out-of-fold response values which is captured by the out-of-fold
predictions:
\begin{equation*}
  \textrm{pR}^2 = 1 - \frac{\sum (y-\hat{y})^2)}{\sum (y - \bar{y})^2)}.
\end{equation*}

\begin{figure}
  \begin{center}
    \includegraphics[width=\textwidth]{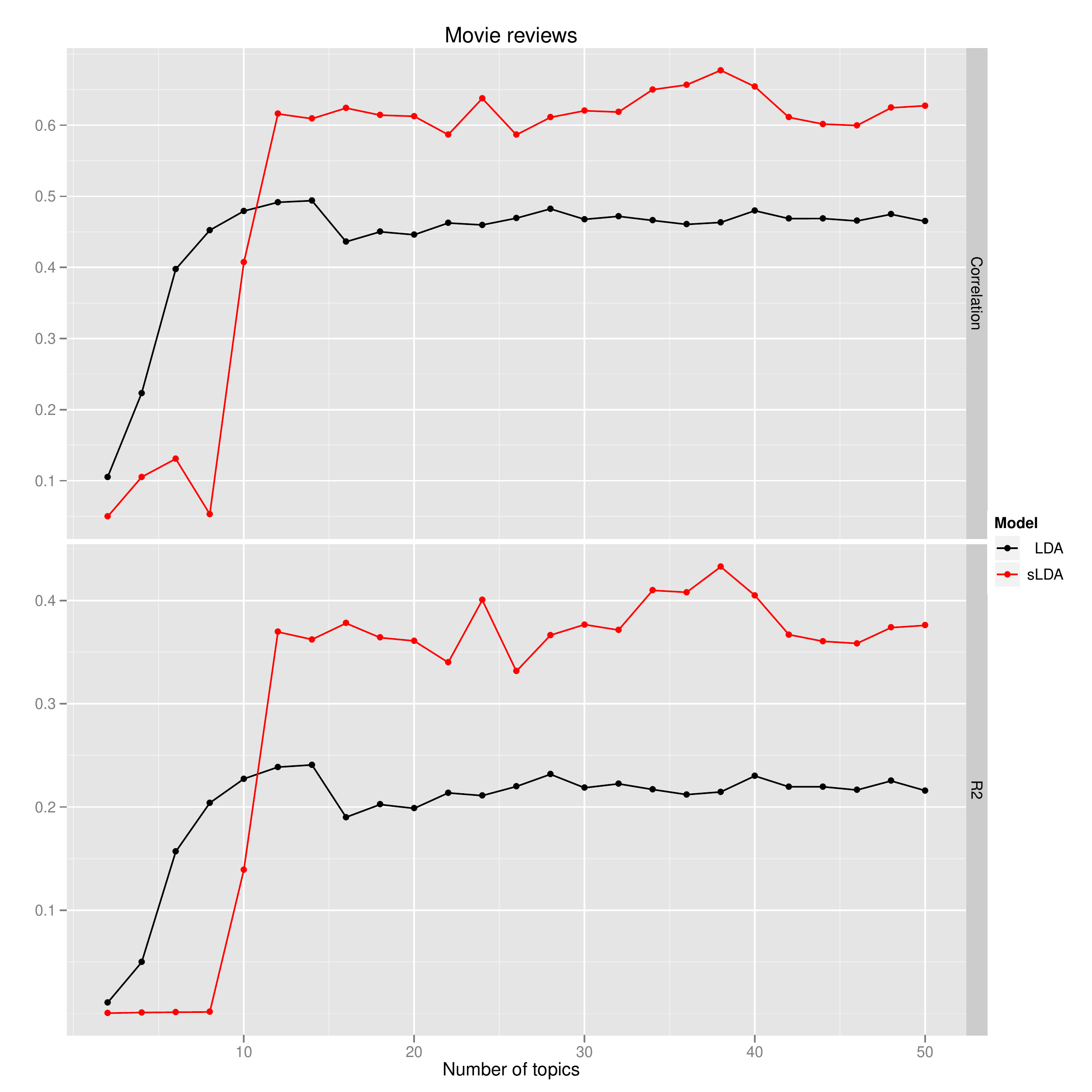}
  \end{center}
  \caption{Error between out-of-fold predictions and observed response
    on the movie review data.\label{fig:review-results}}
\end{figure}

\begin{figure}
  \begin{center}
    \includegraphics[width=\textwidth]{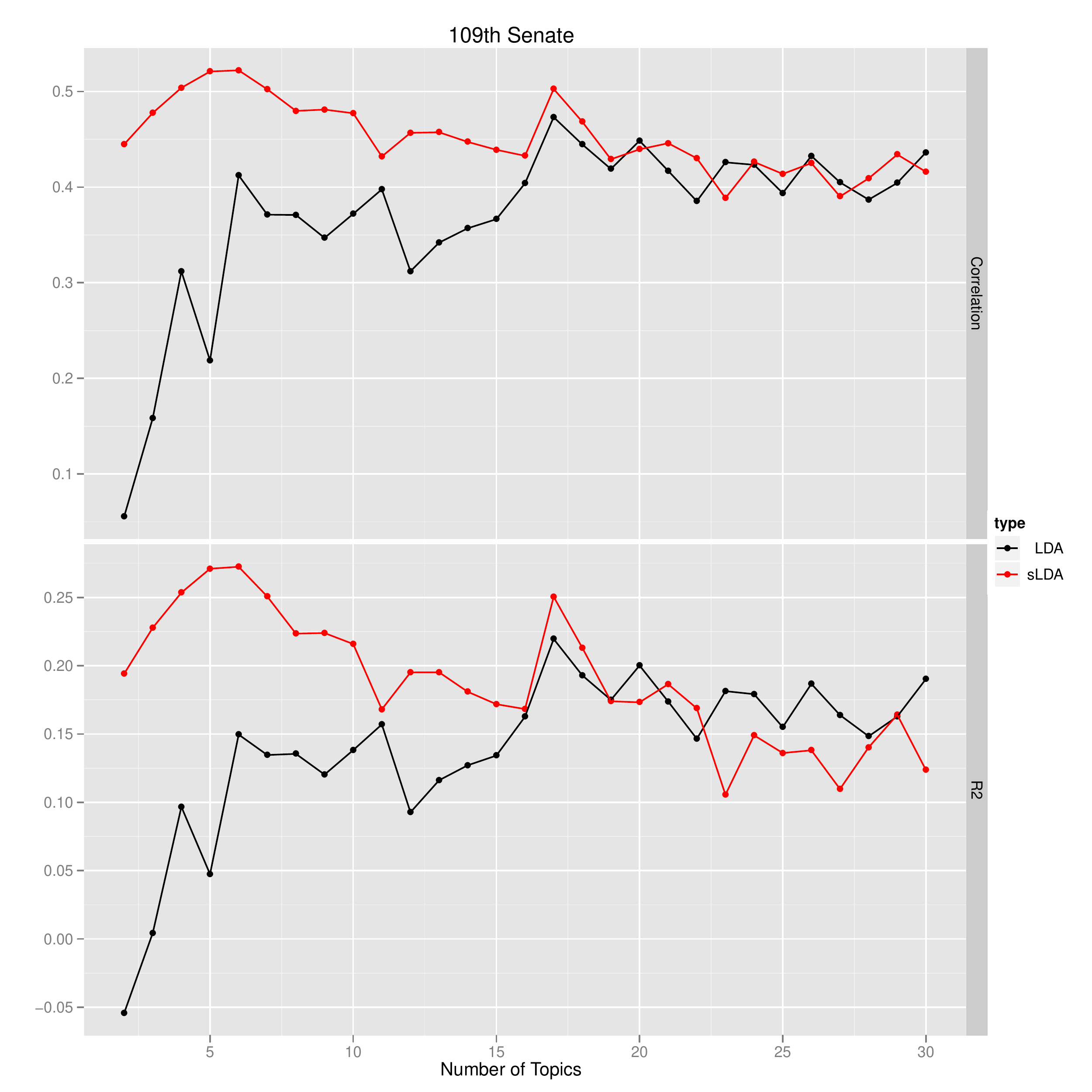}
  \end{center}
  \caption{Error between out-of-fold predictions and observed response
    on the 109th U.S. Senate data.\label{fig:s109-results}}
\end{figure}

\begin{figure}
  \begin{center}
    \includegraphics[width=\textwidth]{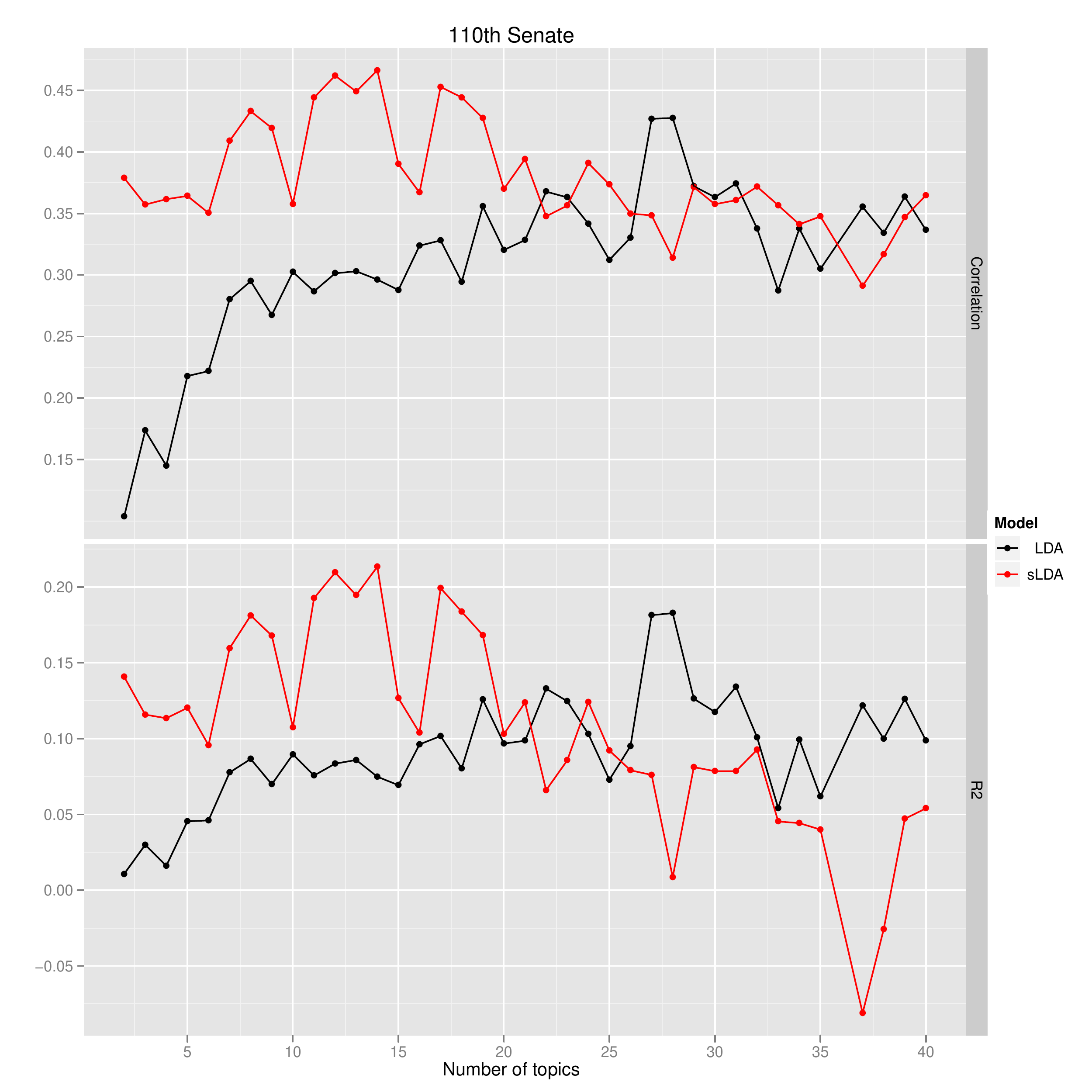}
  \end{center}
  \caption{Error between out-of-fold predictions and observed response
    on the 110th U.S. Senate data.\label{fig:s110-results}}
\end{figure}

We compared sLDA to linear regression on the $\bar{\phi}_d$ from
unsupervised LDA. This is the regression equivalent of using LDA
topics as classification features~\citep{Blei:2003b, Fei-Fei:2005}.
We studied the quality of these predictions across different numbers
of topics.  Figures \ref{fig:review-results}, \ref{fig:s109-results}
and \ref{fig:s110-results} illustrate that sLDA provides improved
predictions on all data.  The movie review rating is easier to predict
that the U.S. Senates discrimination parameters.  However, our study
shows that there is predictive power in the texts of the amendments.

Finally, we compared sLDA to the lasso, which is L$_1$-regularized
least-squares regression. The lasso is a widely used prediction method
for high-dimensional problems~\citep{Tibshirani:1996}. We used each
document's empirical distribution over words as its lasso covariates.
We report the highest $\prr$ it achieved across different settings of
the complexity parameter, and compare to the highest pR$^2$ attained
by sLDA across different numbers of topics.  For the reviewer data,
the best lasso $\prr$ was $0.426$ versus $0.432$ for sLDA, a modest
$2\%$ improvement.  On the U.S. Senate data, sLDA provided
definitively better predictions.  For the 109th U.S. Senate data the
best lasso $\prr$ was $0.15$ versus $0.27$ for sLDA, an $80\%$
improvement.  For the 110th U.S. Senate data, the best lasso $\prr$
was $0.16$ versus $0.23$ for sLDA, a $43\%$ improvement.  Note
moreover that the Lasso provides only a prediction rule, whereas sLDA
models latent structure useful for other purposes.

\section{Discussion}

We have developed sLDA, a statistical model of labelled documents. The
model accommodates the different types of response variable commonly
encountered in practice. We presented a variational procedure for
approximate posterior inference, which we then incorporated in an EM
algorithm for maximum-likelihood parameter estimation. We studied the
model's predictive performance, our main focus, on two real-world
problems.  In both cases, we found that sLDA improved on two natural
competitors: unsupervised LDA analysis followed by linear regression,
and the lasso.  These results illustrate the benefits of supervised
dimension reduction when prediction is the ultimate goal.

We close with remarks on future directions. First, a
``semi-supervised'' version of sLDA---where some documents have a
response and others do not---is straightforward.  Simply omit the last
two terms in~\myeq{phi-gradient} for unlabelled documents and include
only labelled documents in~\myeq{glm-eta-gradient}
and~\myeq{delta-deriv}. Since partially labelled corpora are the rule
rather than the exception, this is a valuable avenue.  (Though note,
in this setting, that care must be taken that the response data exert
sufficient influence on the fit.)  Second, if we observe an additional
fixed-dimensional covariate vector with each document, we can include
it in the analysis just by adding it to the linear predictor. This
change will generally require us to add an intercept term as
well. Third, the technique we have used to incorporate a response can
be applied in existing variants of LDA, such as author-topic
models~\citep{Rosen-Zvi:2004}, population-genetics
models~\citep{Pritchard:2000}, and survey-data
models~\citep{Erosheva:2002}.  We have already mentioned that sLDA has
been adapted to network models~\citep{Chang:2010} and image
models~\citep{Wang:2009a}.  Finally, we are now studying the
maximization of conditional likelihood rather than joint likelihood,
as well as Bayesian nonparametric methods to explicitly incorporate
uncertainty about the number of topics.

\bibliography{slda}

\end{document}